\pdfoutput=1
\documentclass[10pt,final,conference,letterpaper]{ieeeconf}
\IEEEoverridecommandlockouts 
\usepackage{amsmath}

\DeclareMathOperator*{\argmax}{arg\,max}
\usepackage{amsfonts}
\usepackage{pgf}
\usepackage{tikz}
\usepackage{bm}
\usepackage{enumerate}
\usepackage[T1]{fontenc}
\usepackage[noadjust]{cite}
\usepackage{subcaption}
\usepackage{algorithmicx}
\usepackage[ruled, vlined, plain,linesnumbered]{algorithm2e}

\usepackage{fancyhdr}

\title{\LARGE \bf Driving with Style: Inverse Reinforcement Learning in General-Purpose Planning for Automated Driving}
\author{Sascha Rosbach$^{1,2}$, Vinit James$^1$, Simon Gro{\ss}johann$^1$, Silviu Homoceanu$^1$ and Stefan Roth$^2$
\thanks{$^{1}$The authors are with the Volkswagen AG, 38440 Wolfsburg, Germany
        {\tt\small \{sascha.rosbach, vinit.james, simon.grossjohann, silviu.homoceanu\}@volkswagen.de}}%
\thanks{$^{2}$The authors are with the Visual Inference Lab,
        Department of Computer Science, Technische Universit\"at Darmstadt,
        64289 Darmstadt
        {\tt\small stefan.roth@visinf.tu-darmstadt.de}}%
}

\begin{document}
\maketitle
\thispagestyle{fancy} 
\begin{abstract}
Behavior and motion planning play an important role in automated driving.
Traditionally, behavior planners instruct local motion planners with predefined behaviors.
Due to the high scene complexity in urban environments, unpredictable situations may occur in which behavior planners fail to match predefined behavior templates.
Recently, general-purpose planners have been introduced, combining behavior and local motion planning.
These general-purpose planners allow behavior-aware motion planning given a single reward function.
However, two challenges arise:
First, this function has to map a complex feature space into rewards.
Second, the reward function has to be manually tuned by an expert.
Manually tuning this reward function becomes a tedious task.
In this paper, we propose an approach that relies on human driving demonstrations to automatically tune reward functions.
This study offers important insights into the driving style optimization of general-purpose planners with maximum entropy inverse reinforcement learning.
We evaluate our approach based on the expected value difference between learned and demonstrated policies.
Furthermore, we compare the similarity of human driven trajectories with optimal policies of our planner under learned and expert-tuned reward functions.
Our experiments show that we are able to learn reward functions exceeding the level of manual expert tuning without prior domain knowledge.
\end{abstract}
\section{Introduction}
The trajectory planner in highly automated vehicles must be able to generate comfortable and safe trajectories in all traffic situations.
As a consequence, the planner must avoid collisions, monitor traffic rules, and minimize the risk of unexpected events.
General-purpose planners fulfill these functional requirements by optimization of a complex reward function.
However, the specification of such a reward function involves tedious manual tuning by motion planning experts.
Tuning is especially tedious if the reward function has to encode a humanlike driving style for all possible scenarios.
In this paper, we are concerned with the automation of the reward function tuning process.

\begin{figure}  
\vspace{1.5mm}
\centering
\includegraphics[width=0.48\textwidth]{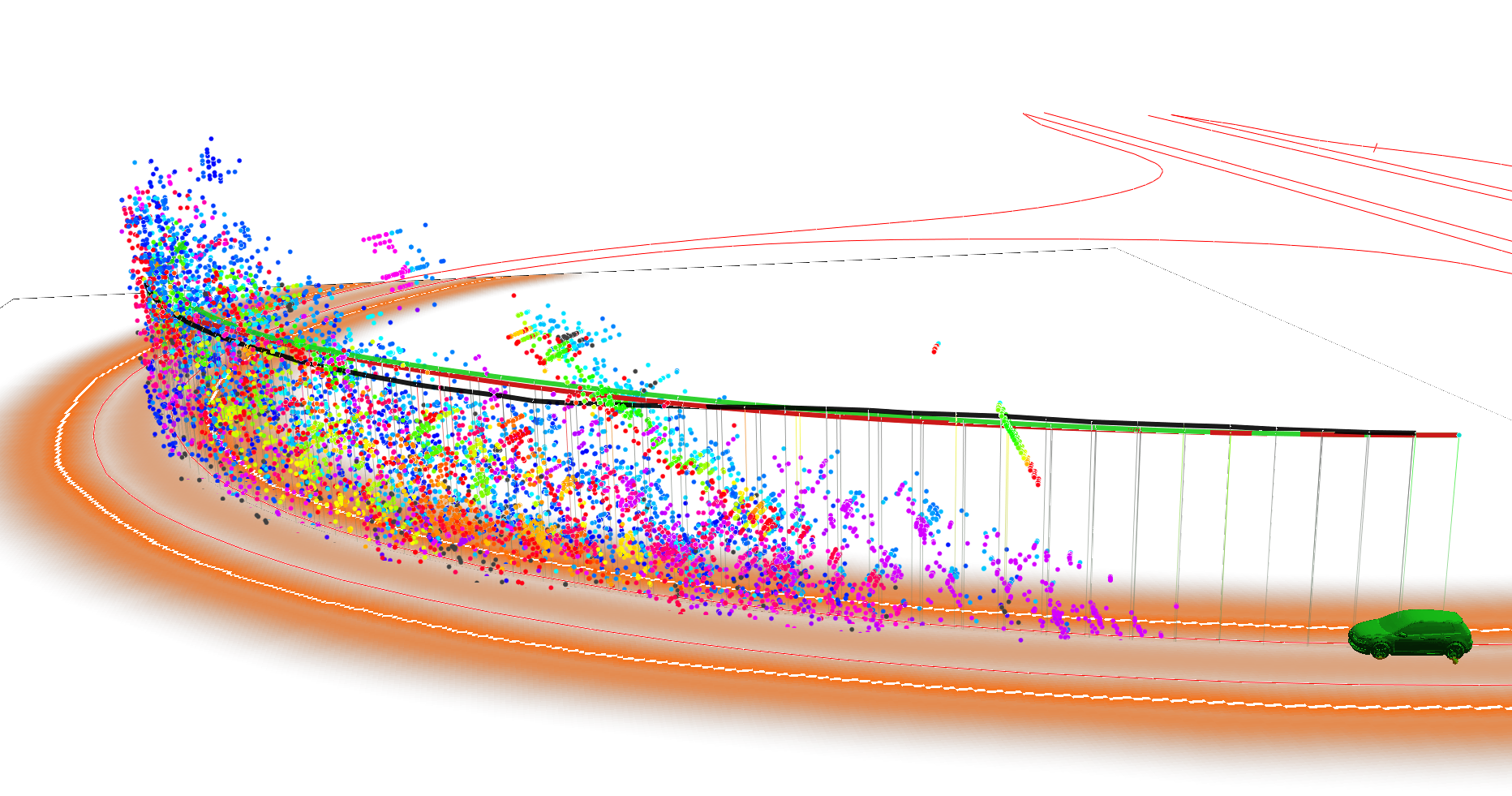}
\caption{
  This figure illustrates our general-purpose planner for automated driving.
  The color coding of the visualized state space indicates the state-action values.
  The z-axis corresponds to the velocity, while the groundplane depicts a subset of spatial features such as distance transformed lane centers and road boundaries.
  There are three color coded policies, black denotes the optimal policy of the planner, red the odometry of a human demonstration, and green the projection of the demonstration into the state space.
  }
\label{fig:title}
\end{figure}

Unlike a strict hierarchical planning system, our planner integrates behavior and local motion planning.
The integration is achieved by a high-resolution sampling with continuous actions~\cite{mcnaughton2011phd}.
Our planner, shown in Fig.~\ref{fig:title}, derives its actions from a vehicle transition model.
This model is used to integrate features of the environment, which are then used to formulate a linear reward function.
During every planning cycle of a model predictive control (MPC), the planning algorithm generates a graph representation of the high-dimensional state space. 
At the end of every planning cycle, the algorithm yields a large set of driving policies with multiple implicit behaviors, e.g., lane following, lane changes, swerving and emergency stops.
The final driving policy has the highest reward value while satisfying model-based constraints.
The reward function, therefore, influences the driving style of all policies without compromising safety.

Human driving demonstrations enable the application of inverse reinforcement learning (IRL) for finding the underlying reward functions, i.e., a linear combination of the reward weights.
In this work, we utilize this methodology to automate the reward function tuning of our planner.
Due to the planner's exploration of a large set of actions, we are able to project demonstrated actions into our graph representation.
Thereby, the demonstrations and associated features are efficiently captured.
As a result, the learning algorithm enables the imitation of the demonstrated driving style.
Most related work in IRL utilizes the state visitation frequency to calculate the gradient in \textit{maximum entropy} IRL.
However, the calculation of the state visitation is generally intractable in this high-dimensional state space.
We utilize our graph representation to approximate the required empirical feature expectations to allow \textit{maximum entropy} IRL.

The main contributions of this paper are threefold:
First, we formulate an IRL approach which integrates maximum entropy IRL with a model-predictive general-purpose planner.
This formulation allows us to encode a humanlike driving style in a linear reward function.
Second, we demonstrate the superiority of our automated reward learning approach over manual reward tuning by motion planning experts.
We draw this conclusion on the basis of comparisons over various performance metrics as well as real world tests.
Third, our automated tuning process allows us to generate multiple reward functions that are optimized for different driving environments and thereby extends the generalization capability of a linear reward function.

\section{Related Work}

The majority of active planning systems for automated driving are built on the mediated perception paradigm.
This paradigm divides the automated driving architecture into sub-systems to create abstractions from raw sensory data.
The general architecture includes a perception module, a system to predict the intention of other traffic participants, and a trajectory planning system.
The planning system is usually decomposed in a hierarchical structure to reduce the complexity of the decision making task~\cite{katrakazas2015,paden2016, schwarting2018}.
On a strategic level, a route planning module provides navigational information.
On a tactic and behavioral level, a behavioral planner derives the maneuver, e.g., lane-change, lane following, and emergency-breaking~\cite{ulbrich2015itsc}.
On an operational level, a local motion planner provides a reference trajectory for feedback control~\cite{werling2010icra}.
However, these hierarchical planning architectures suffer from uncertain behavior planning due to insufficient knowledge about motion constraints.
As a result, a maneuver may either be infeasible due to over-estimation or discarded due to under-estimation of the vehicle capabilities.
Furthermore, behavior planning becomes difficult in complex and unforeseen driving situations in which the behavior fails to match predefined admissibility templates.
Starting with the work of McNaughton, attention has been drawn to parallel real-time planning~\cite{mcnaughton2011phd}.
This approach enables sampling of a large set of actions that respect kinematic constraints.
Thereby a sequence of sampled actions can represent complex maneuvers.
This kind of general-purpose planner uses a single reward function, which can be adapted online by a behavioral planner without the drawbacks of a hierarchical approach.
However, it is tedious to manually specify and maintain a set of tuned reward functions. 
The process required to compose and tune reward functions is outside the scope of McNaughton's work. 
We adopt the general-purpose planning paradigm in our approach and focus on the required tuning process.

Reward functions play an essential role in general-purpose planning.
The rewards encode the driving style and influence the policy selection.
Recently, literature has been published on the feature space of these reward functions.
Heinrich et al.~\cite{heinrich2016iv} propose a model-based strategy to include sensor coverage of the relevant environment to optimize the vehicle's future pose.
Gu et al.~\cite{gu2016iros} derive tactical features from the large set of sampled policies.
So far, however, there has been little discussion about the automated reward function tuning process of a general-purpose planner. 

Previous work has investigated the utilization of machine learning in hierarchical planning approaches to predict tactical behaviors \cite{ulbrich2015itsc}.
Aside of behavior prediction, a large and growing body of literature focuses on finding rewards for behavior planning in hierarchical architectures \cite{abbeel2008iros}, rewards associated with spatial traversability~\cite{wulfmeier2017ijrr}, and rewards for single-task behavior optimization of local trajectory planners \cite{kuderer2015icra}.
The IRL approach plays an import role in finding the underlying reward function of human demonstrations for trajectory planning~\cite{arora2018}.
Similar to this work, several studies have investigated IRL in high-dimensional planning problems with long planning horizons.
Shiarlis et al.~\cite{shiarlis2016aams} demonstrate maximum margin IRL within a randomly-exploring random tree (RRT*).
Byravan et al.~\cite{byravan2015ijcai} focus on a graph-based planning representation for robot manipulation, similar to our planning problem formulation.
Compared to previous work in IRL, our approach integrates IRL directly into the graph construction and allows the application of \textit{maximum entropy} IRL for long planning horizons \textit{without} increasing the planning cycle time.

Compared to supervised learning approaches such as direct imitation and reward learning, reinforcement learning solves the planning problem through learning by experience and interaction with the environment.
Benefits of reinforcement learning are especially notable in the presence of many traffic participants.
Intention prediction of other traffic participants can be directly learned by multi-agent interactions.
Learned behavior may include complex negotiation of multiple driving participants~\cite{shalev-shwartz2016nips}.
Much of the current literature focuses on simulated driving experience and faces challenges moving from simulation to real-world driving, especially in urban scenarios.
Another challenge includes the formulation of functional safety within this approach.
Shalev-Shwartz et al.~\cite{shalev-shwartz2017} describe a safe reinforcement learning approach that uses a hierarchical options graph for decision making where each node within the graph implements a policy function.
In this approach, driving policies are learned whereas trajectory planning is not learned and bound by hard constraints. 

Most of the current work in IRL utilizes the \textit{maximum entropy} principle by Ziebart et al.~\cite{ziebart2008aaai} that allows training of a probabilistic model by gradient descent.
The gradient calculation depends on the state visitation frequency, which is often calculated by an algorithm similar to backward value iteration in reinforcement learning.
Due to the curse of dimensionality, this algorithm is intractable for driving style optimization in high-dimensional continuous spaces.
Our work extends previous work by embedding IRL into a general-purpose planner with an efficient graph representation of the state space.
The design of the planner effectively enables the driving style imitation without a learning task decomposition.
As a result, we utilize the benefits of a model-based general-purpose planner and reward learning to achieve nuanced driving style adaptations.

\section{Preliminaries}
The interaction of the agent with the environment is often formulated as a Markov Decision Process (MDP) consisting of a 5-tuple \{$\mathcal{S}, \mathcal{A},T, R, \gamma$\}, where $\mathcal{S}$ denotes the set of states, and $\mathcal{A}$ describes the set of actions.
A continuous action $a$ is integrated over time $t$ using the transition function $T(s,a,s')$ for ${s, s' \in \mathcal{S}, a \in \mathcal{A}}$.
The reward function $R$ assigns a reward to every action $\mathcal{A}$ in state $\mathcal{S}$.
The reward is discounted by $\gamma$ over time $t$.

In this work, a model of the environment $M$ returns a feature vector $\bm{f}$ and the resultant state $s'$ after the execution of action $a$ in state $s$.
The reward function $R$ is given by a linear combination of $K$ feature values $f_i$ with weights $\theta_i$ such that ${\forall(s,a) \in \mathcal{S} \times \mathcal{A}: R(s,a)=\sum_{i \in K} {-\theta_i f_i{(s,a)}}}$.
A policy $\pi$ is a sequence of time-continuous transitions $T$.
The feature path integral $f^{\pi}_i$ for a policy $\pi$ is defined by ${f_i^{\pi}=\int_t\gamma_t f_i(s_t,a_t)\,dt}$.
The path integral is approximated by the iterative execution of sampled state-action sets $\mathcal{A}_s$ in the environment model $M$.
The value $V^{\pi}$ of a policy $\pi$ is the integral of discounted rewards during continuous transitions $V^{\pi}=\int_t \gamma_t R(s_t,a_t)\,dt$.
An optimal policy $\pi^*$ has maximum cumulative value, where ${\pi^*=\argmax_{\pi} V^{\pi}}$.
A human demonstration $\zeta$ is given by a vehicle odometry record.
A projection of the odometry record $\zeta$ into the state-action space allows us to formulate a demonstration as policy $\pi^D$.
For every planning cycle, we consider a set of demonstrations $\Pi^D$, which are geometrically close to the odometry record $\zeta$.
The planning algorithm returns a finite set of policies $\Pi$ with different driving characteristics.
The final driving policy $\pi^S$ is selected and satisfies model-based constraints.

\section{Methodology}

\begin{figure*}  
 \vspace{1.5mm}
\centering
\includegraphics{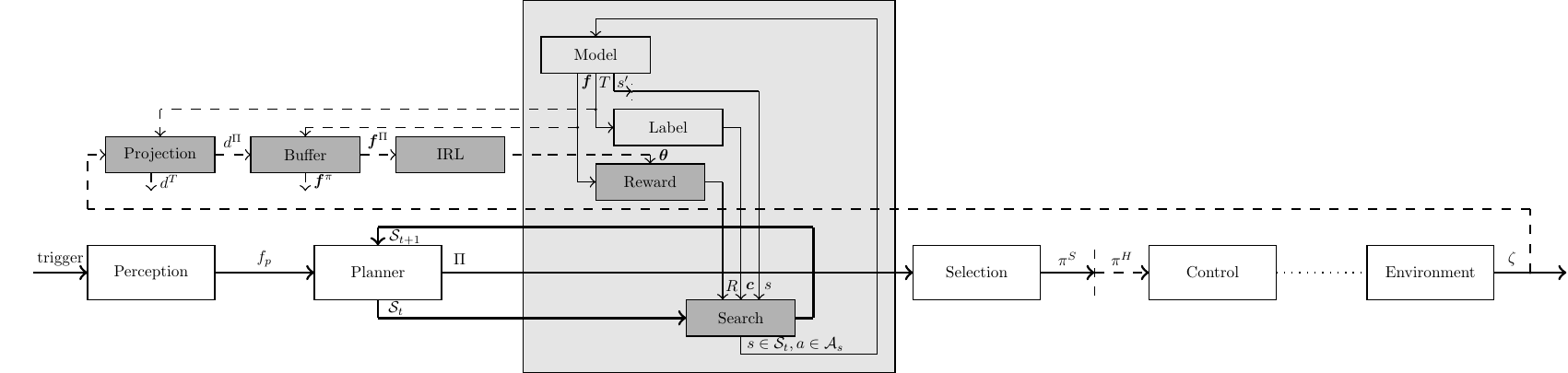}
\caption{
  Functional flow block diagram:
  The left input of a block corresponds to the output of the previous function.
  The inputs on top of the blocks denote intermediate outputs of previous functions. 
  A thick solid line indicates the main flow from the environment perception $f_p$ to the driven trajectory $\zeta$.
  The vehicle control architecture is outside the scope of this work.
  In this work, we focus on the dark grey blocks of the architecture that influence the reward learning.
  Dashed connections between the blocks indicate the information flow during the training procedure.
  During data collection, we record the environment as well as the odometry $\zeta$ of the hidden driving policy of a human $\pi^H$.
  }
\label{fig:architecture}
\end{figure*}

The planning system in Fig.~\ref{fig:architecture} uses MPC to address continuous updates of the environment model.
A periodic trigger initiates the perception system for which the planner returns a selected policy.
In the following, we give an overview of the general-purpose planner.
We use the nomenclature of reinforcement learning to underline the influence of reward learning in the context of search-based planning.
Furthermore, we propose a \textit{path integral maximum entropy} IRL formulation for high-dimensional reward optimization. 

\begin{figure}[t]
  \vspace{1mm}
\includegraphics[trim=0 0 100 0,clip]{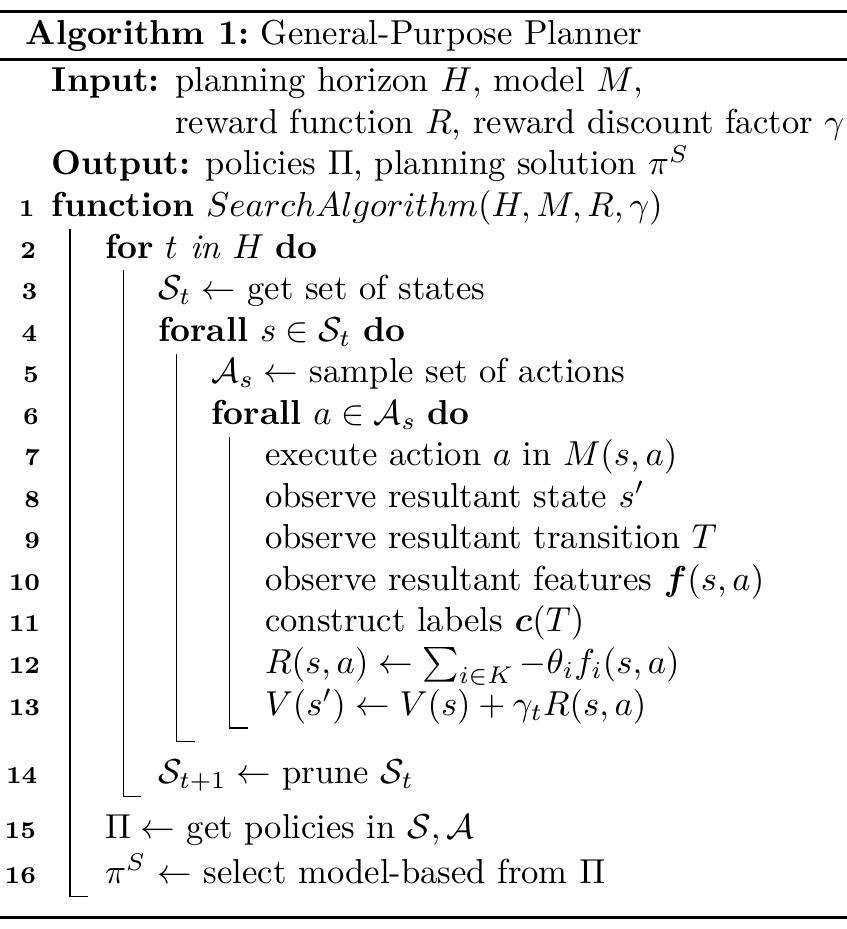}
\end{figure}

\subsection{General-Purpose Planner for Automated Driving}

Our planning algorithm for automated driving in all driving situations is based on~\cite{heinrich2018phd,heinrich2016iv}.
The planner is initialized at state $s_0$, either by the environment model or in subsequent plans by the previous policy, and designed to perform an exhaustive forward search of actions to yield a set of policies $\Pi$.
The set $\Pi$ implicitly includes multiple behaviors, e.g., lane following, lane changes, swerving, and emergency stops~\cite{mcnaughton2011phd}.
Fig.~\ref{fig:architecture} visualizes the functional flow of the planning architecture during inference and training. 

Algo.~1 formally describes our search-based planning approach.
The planner generates trajectories for a specified planning horizon $H$.
Trajectories for the time horizon $H$ are iteratively constructed by planning for discrete transition lengths.
The planner uses the parallelism of a graphics processing unit (GPU) to sample for all states $s \in \mathcal{S}_t$ a discrete number of continuous actions $\mathcal{A}_s$, composed of accelerations and wheel angles.
The sampling distribution for each state is based on feasible vehicle dynamics.
The actions itself are represented by time-continuous polynomial functions, where order and coefficients are derived from actor-friendly continuity constraints.
This results in longitudinal actions described by velocity profiles up to fifth order, and lateral actions described by wheel angle profiles up to third order.

The search algorithm calls the model of the environment $M$ for all states $s \in \mathcal{S}_t$ to observe the resultant state $s'$, transition $T$, and features $\bm{f}$ for each state-action tuple.
The feature vector $\bm{f}$ is generated by integrating the time-continuous actions in the environment model.
A labelling function assigns categorical labels to transitions, e.g., a label associated with collision.
A pruning operation limits the set of states $\mathcal{S}_{t+1}$ for the next transition step $t+1 \in H$.
Pruning is performed based on the value $V(s)$, label $c$, and properties of the reachable set $\mathcal{S}_t$ to terminate redundant states with low value $V(s)$.
This operation is required, first to limit the exponential growth of the state space, and second to yield a policy set $\Pi$ with maximum behavior diversity.
The algorithm is similar to parallel breadth first search and forward value iteration. 
The final driving policy $\pi^S$ is selected based on the policy value $V(\pi)$ and model-based constraints.

\subsection{Inverse Reinforcement Learning}
The driving style of a general-purpose motion planner is directly influenced by the reward function weights $\bm{\theta}$.
The goal of IRL is to find these reward function weights $\bm{\theta}$ that enable the optimal policy $\pi^*$ to be at least as good as the demonstrated policy $\pi^D$, i.e., $V(\pi^*) \geq V(\pi^D) $.
Thereby, the planner indirectly imitates the behavior of a demonstration~\cite{ng2000icml}.
However, learning a reward function given an optimal policy is ambiguous since many reward functions may lead to the same optimal policy~\cite{abbeel2004icml}.
Early work in reward learning for A* planning and dynamic programming approaches utilized structured maximum-margin classification~{\cite{ratliff2006icml}, yet this approach suffers from drawbacks in the case of imperfect demonstrations~\cite{ziebart2008aaai}.
Over the past decade, most research in IRL has focused on maximizing the entropy of the distribution on state-actions under the learned policy, which is known as \textit{maximum entropy} IRL.
This problem formulation solves the ambiguity of imperfect demonstrations by recovering a distribution over potential reward functions while avoiding any bias~\cite{ziebart2008aaai}.
Ziebart et al.~\cite{ziebart2008aaai} propose a state visitation calculation, similar to backward value iteration in reinforcement learning, to compute the gradient of the entropy.
The gradient calculation is adopted by most of the recent work in IRL for low-dimensional, discrete action spaces, which is inadequate for driving style optimizations.
Our desired driving style requires high-resolution sampling of time-continuous actions, which produces a high-dimensional state space representation.
In the following, we describe our intuitive approach, which combines search-based planning with \textit{maximum entropy} IRL.

\subsection{Path Integral Maximum Entropy IRL}

In our IRL formulation, we maximize the log-likelihood $L$ of expert behavior in the policy set $\Pi$ by finding the reward function weights $\bm{\theta}$ that best describe human demonstrations $\pi^D \in \Pi^D$ within a planning cycle, which is given by
  \begin{align}
  \bm{\theta}^* &= \argmax_{\bm{\theta}} L(\bm{\theta}) = \argmax_{\bm{\theta}}\sum_{\pi^D \in \Pi^D}\ln{p(\pi^D|\bm{\theta})}\\
    &= \argmax_{\bm{\theta}}\sum_{\pi^D \in \Pi^D}\ln\dfrac{1}{Z}\exp(\bm{-\theta f}^{\pi^D}),
  \label{eq:likelihood}
  \end{align}
where the partition function is defined by ${Z=\sum_{\pi \in \Pi}{\mbox{exp}(\bm{-\theta f}^{\pi})}}$.

Similar to Aghasadeghi et al. \cite{aghasadeghi2011iros}, we optimize under the constraint of matching the feature path integrals $\bm{f}^\pi$ of the demonstration and feature expectations of the explored policies, 

\begin{equation}
  \forall i \in {1,...,k}: \sum_{\pi \in \Pi} p(\pi|\theta)f_i^\pi =\dfrac{1}{m} \sum_{\pi^D \in \Pi^D}  f^{\pi^D}_i = \hat{f}^{\Pi^D}_i,
  \label{eq:constraint}
\end{equation}
where $\hat{f}^{\Pi^D}_i$ references the empirical mean of feature $i$ calculated over $m$ demonstrations in $\Pi^D$.
The constraint in Eq. \ref{eq:constraint} is used to solve the non-linear optimization in Eq. \ref{eq:likelihood}.

The gradient of the log-likelihood can be derived as,
\begin{equation}
 \nabla L(\bm{\theta}) = \sum_{\pi \in \Pi} p(\pi|\bm{\theta})\bm{f}^\pi - \bm{\hat{f}}^{\Pi^D}, 
\end{equation}
and allows for gradient descent optimization.

The calculation of the partition function $Z$ in Eq. \ref{eq:likelihood} is often intractable due to the exponential growth of the state-action space over the planning horizon.
The parallelism of the action sampling of the search-based planner allows us to explore a high-resolution state representation $S_t$ for each discrete planning horizon increment $t$.
A pruning operation terminates redundant states having sub-optimal behavior in the reachable set $S_t$, which is denoted by a lower value $V(s)$. 
Therefore, the pruning operation ensures multi-behavior exploration of the reachable set $S_t$ that is evaluated with a single reward function.
Thereby our sample-based planning methodology allows us to approximate the partition function similar to Markov chain Monte Carlo methods.

Once we obtain the new reward function, the configuration of the planner is updated.
Hence, policies that have similar features as the human demonstration acquire a higher value assignment.
This implies that they are more likely to be chosen as driving policy.

\section{Experiments}

We assess the performance of \textit{path integral maximum entropy } IRL in urban automated driving.
We focus on a base feature set for static environments, similar to the manual tuning process of a motion planning expert.
After this process more abstract reward features are tuned relative to the base features.

\subsection{Data Collection and Simulation}

Our experiments are conducted on a prototype vehicle, which uses a mediated perception architecture to produce feature maps as illustrated in Fig. \ref{fig:title}.
We recorded data in static environments and disabled object recognition and intention prediction.
The data recordings include features of the perception system as well as odometry recordings of the human driver's actions.
The training of our algorithm is performed during playbacks of the recorded data.
After every planning cycle of the MPC, the position of the vehicle is reset to the odometry recording of the human demonstration.

\subsection{Projection of Demonstration in State Space}

The system overview in Fig. \ref{fig:architecture} includes a projection function that transfers the actions of a manual drive into the state-action space of the planning algorithm.
The projection metric $d$ is calculated during the graph construction between odometry $\zeta$ and continuous transitions $T(s,a, s')$ of all policies $\pi$ in the set $\Pi$:
\begin{equation}
  d(\zeta,\pi) = \int\limits_t \alpha_t || \zeta_{t} - \pi_{t}||\,dt.
  \label{eq:distance}
\end{equation}

The norm is based on geometrical properties of the state space, e.g., the Euclidean distance in longitudinal and lateral direction as well as the squared difference in the yaw angle.
Further, the metric includes a discount factor $\alpha_t$ over the planning horizon.
The policy $\pi^D$ has the least discounted distance towards the odometry record.
There are multiple benefits of using the projection metric:
First, the projected trajectory includes all constraints of the planner.
If the metric surpasses a threshold limit, the human demonstrator does not operate in the actor's limits of the vehicle and therefore can not be used as a valid demonstration.
Second, the projection metric allows for an intuitive evaluation of the driving style based on the geometrical proximity to the odometry.
Third, we may augment the number of demonstrations by loosening the constraint of the policy $\pi^D$ to have least discounted distance towards the odometry.
Thereby, multiple planner policies qualify as demonstration $\pi^D\,\subseteq\,\Pi^D$.

\subsection{Reward Feature Representation}

In this work, the reward function $R(s,a)$ is given by a linear combination of $K$ reward features.
The features describe motion and infrastructural rewards of driving. 
The discount factor $\gamma$ is manually defined by a motion planning expert and is not optimized at this stage of the work.
Our perception system in Fig.~\ref{fig:architecture} provides normalized feature maps with spatial information of the environment.
The feature path integral $\bm{f}^{\pi}$ of a policy $\pi$ is created by transitioning through the feature map representation of the environment. 
We concentrate on a base reward set consisting of $K=12$ features, which are listed in the legend of Fig.~\ref{fig:allweights}.
Heinrich et al. formally described a sub-set of our features~\cite{heinrich2018phd}.
Seven of our feature values describe the motion characteristics of the policies, which are given by derivatives of the lateral and longitudinal actions.
They include the difference between the target and policy velocity, and the acceleration and jerk values of actions.
The target in the velocity may change depending on the situation, e.g., in a situation with a traffic light the target velocity may reduce to zero.
Furthermore, the end direction feature is an important attribute for lateral behavior that specifies the angle towards the driving direction at the end of the policy. 
The creeping feature suppresses very slow longitudinal movement in situations, where a full stop is more desired.
Infrastructural features include proximity measures to the lane center and curbs, cost potentials for lanes, and direction.
Furthermore, we specify a feature for conflict areas, e.g., stopping at a zebra crossing.

\subsection{Implementation Details}

During the playback of a human demonstration, the path integral feature vectors $\bm{f}^{\Pi}$ of the policy set $\Pi$ are approximated for every planning cycle and stored within a replay buffer.
By including our projection metric in the action sampling procedure, we associate each policy $\pi$ with the distance to the odometry of the human demonstration.
During training, we query demonstrations $\Pi^D$, which are policies with a low projection metric value, from our replay buffer where $\pi^D\,\subseteq~\Pi^D\,\subseteq\,\Pi$.
Hence, the replay buffer contains features of demonstrations for each planning cycle denoted as $\bm{f}^{\Pi^D}\,\subseteq\,\bm{f}^{\Pi}$.
Fig. \ref{fig:architecture} describes the information flow from the odometry record of the demonstration to the feature query from the replay buffer.
Due to actor constraints of the automated vehicle's actions, the planning cycles without demonstrations are not considered for training.
We utilize experience replay similar to reinforcement learning and update on samples or mini-batches of experience, by drawing randomly from the buffered policies.
This process allows us to efficiently use previous experience, which can be trained on multiple times.
Further, stability is provided by not altering the representation of the expert demonstrations within the graph representation.

\begin{figure*}[h]
  \centering
  \vspace{1mm}
    \begin{subfigure}[t]{0.4\textwidth}
      \centering
      \includegraphics{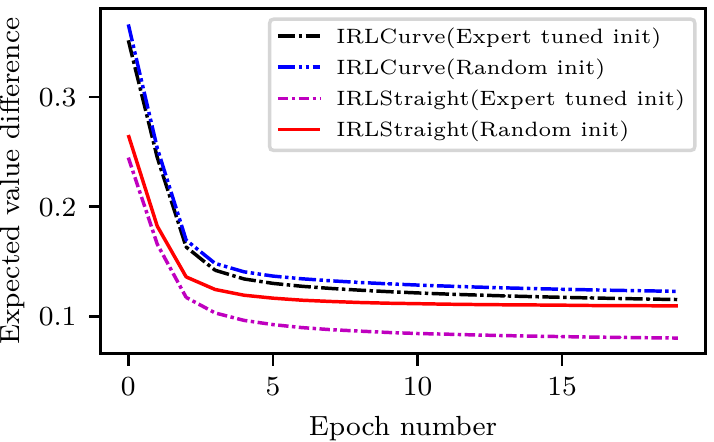}
      \caption{Difference between expected value of human driving demonstration and expected value of planner policies under learned reward functions.}
      \label{fig:evd}
    \end{subfigure}
    \hspace{2mm}
    \begin{subfigure}[t]{0.4\textwidth}
      \centering
      \includegraphics{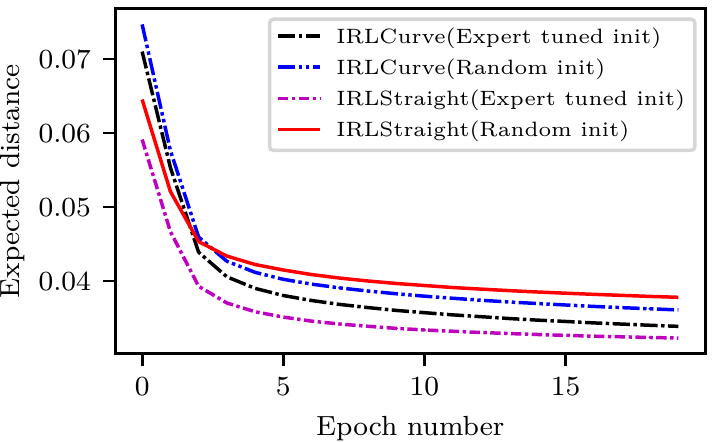}
      \caption{Expected distance of planner policies towards the human driving demonstration under learned reward functions.}
      \label{fig:ed}
	\end{subfigure}
  \caption{
Illustration of training and validation metrics for multiple segments and training initializations.
Convergence of maximum entropy IRL over training epochs. 
Validation of the training by indicating the reduction of the expected distance towards the human demonstration.
The probability is calculated independently for every planning cycle of the MPC, whereas the policy set includes an average of approx. 4000 policies.}
  \label{fig:evd}
\end{figure*}

\section{Evaluation}

We aim to evaluate the utility of our automated reward function optimization in comparison to manual expert tuning.
First, we analyze our driving style imitation in terms of value convergence towards the human demonstration.
Second, we compare the driving style of our policies under random, learned, and expert-tuned reward functions against human driving on dedicated test route segments.

\subsection{Training Evaluation}

We analyze the convergence for different training initializations and road segment types, namely straight and curvy.
Due to the linear combination of reward weights, one expects a segment-specific preference of the reward function.
As a reference, a motion planning expert generated a tuned reward function for general driving.
We perform two drives per training segment, one with a random and one with an expert-tuned reward function.
The policies to be considered as human demonstrations are chosen based on our projection metric and therefore depend on our chosen reward function initialization.
The expert initialization yields demonstrations with a mean projection error 7\% lower as compared to random initialization.
During every planning cycle on the segments, we trace the policies of the planner in replay buffers.
We generate four tuned reward functions which are referred to in Fig.~\ref{fig:allweights} by training on our replay buffers. 

The convergence of the training algorithm is measured by the expected value difference (EVD) over training epochs between learned and demonstrated policies.
The EVD is calculated for every planning cycle and averaged over the segment.
The EVD is given by
\begin{align}
 \mathbb{E}[V(\Pi)]&-\mathbb{E}[V(\Pi^D)] \\
\label{Eq:evd}
 =\sum_{\pi \in \Pi} {p(\pi|\bm{\theta}) V(\pi)}&-\sum_{\pi^D \in \Pi^D} {p(\pi^D|\bm{\theta}) V(\pi^D)}.
\end{align}

The performance of the random and expert-tuned reward functions is given by the EVD at epoch zero.
The initial and final EVD differences between the straight and curvy segment is 30\% and 19\% respectively.
A preference of the straight segments by both reward functions is visible in the initial EVD difference~Fig.~\ref{fig:evd}.
The learned reward functions show a large EVD reduction of 67\% for curvy and 63\% for straight segments at the end of training.

We can interpret the training results in the following way:
\begin{enumerate}[(a)]
\item The projection metric depends on the quality of the reward function.
\item Improved reward functions lead to improved action sampling and therefore produce better demonstrations.
\item Learning reward functions without prior knowledge is possible, e.g. generating a replay buffer with a randomly initialized reward function and training with a random initialization.
\item Unsuitable reward functions improve more significantly during training.
\end{enumerate}
Hence, continuously updating the policies in the replay buffer generated from an updated reward function should lead to faster convergence.

The desired driving style is given by the actions of a human driving demonstration.
Therefore, the projection error in Eq.~\ref{eq:distance}, which we use to select driving demonstrations, extends itself as a direct validation metric of the actions.
Due to our goal of optimizing the likelihood of human driving behavior within the policy set, we calculate the expected distance (ED) in the policy set, given by
\begin{equation}
  \mathbb{E}[d(\zeta,\Pi)]= \sum_{\pi \in \Pi} {p(\pi|\bm{\theta}) d(\zeta,\pi)}.
  \label{eq:ed}
\end{equation}
The learned reward functions in Fig.~\ref{fig:ed} show a large ED reduction of 54\% for curvy and 44\% for straight segments at the end of training.
The ED reduction trends have high similarity to the above mentioned EVD trend and therefore this validates the premise of a high correlation between value and distance to the demonstration.
An improved expected distance ensures a high likelihood of selecting policies which are similar to humanlike driving demonstrations. 

\subsection{Driving Style Evaluation}

\begin{figure*}
  \centering
  \vspace{1mm}
            \begin{subfigure}[t]{0.60\textwidth}
              \includegraphics[trim=0 22 0 0, clip]{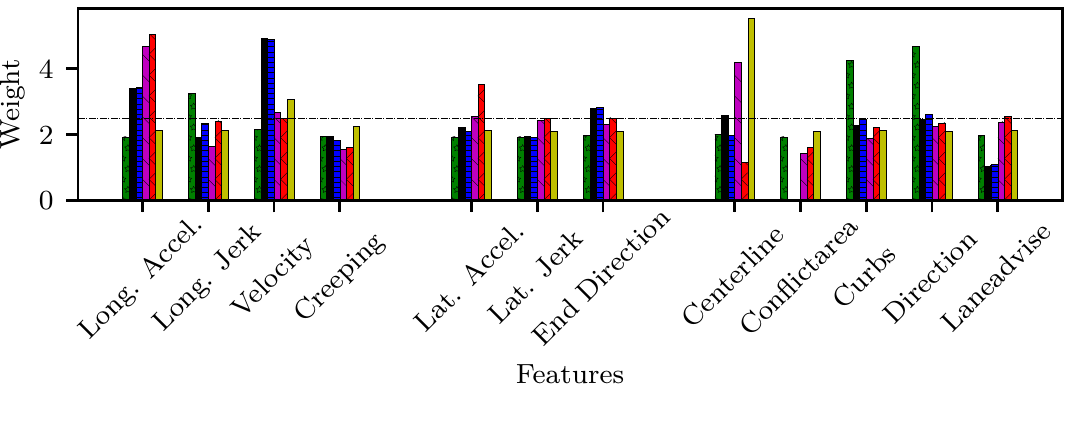}
              \caption{Feature weights of tested reward functions.} 
              \label{fig:allweights}
              \vspace{2mm}
            \end{subfigure}
            \hspace{2mm}
            \begin{subfigure}[t]{0.25\textwidth}
              \centering
              \includegraphics[trim=0 0 0 10, clip]{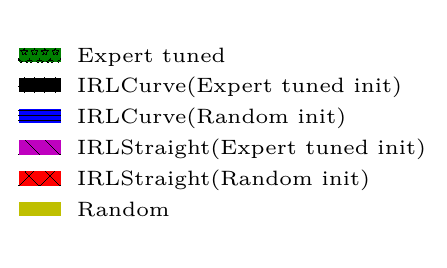}
              \label{fig:testlegend}
            \end{subfigure}
            \begin{subfigure}[t]{0.40\textwidth}
              \includegraphics[trim=0 5 0 0, clip]{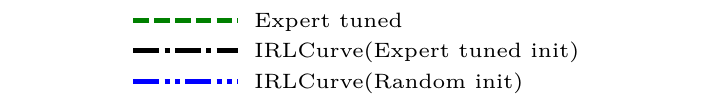}
              \label{fig:legend}
            \end{subfigure}
            \hspace{2mm}
            \begin{subfigure}[t]{0.40\textwidth}
              \includegraphics[trim=0 5 0 0, clip]{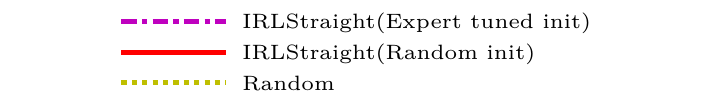}
              \label{fig:legend}
            \end{subfigure}
            \begin{subfigure}[t]{0.40\textwidth}
              \centering
              \includegraphics{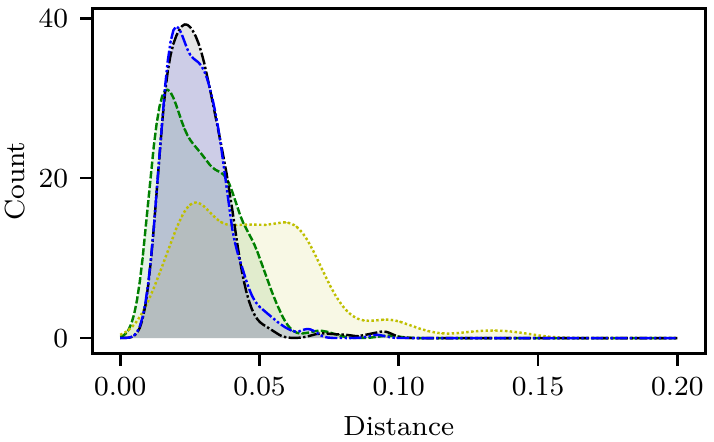}
              \caption{Distances of the optimal policies on a curvy test segment.}
              \label{fig:curvydisthist}
            \end{subfigure}
            \hspace{2mm}
            \begin{subfigure}[t]{0.40\textwidth}
              \centering
              \includegraphics{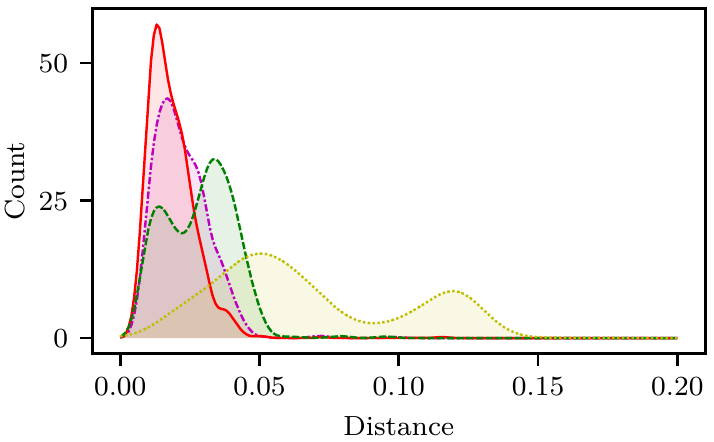}
              \caption{Distances of the optimal policies on a straight test segment.}
              \vspace{2mm}
              \label{fig:straightdisthist}
            \end{subfigure}
            \begin{subfigure}[t]{0.40\textwidth}
              \centering
              \includegraphics{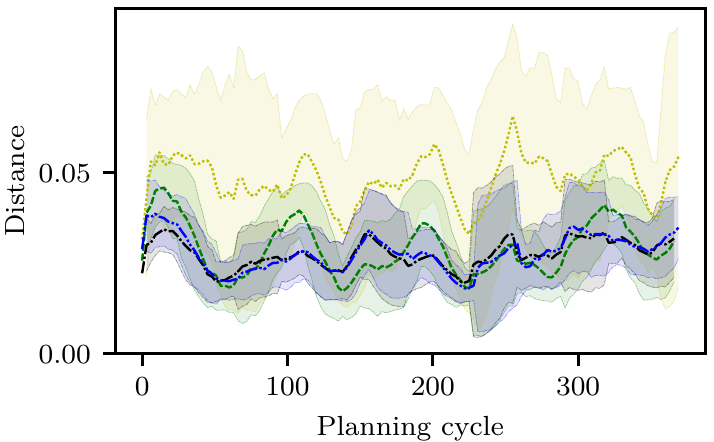}
              \caption{Distances of the optimal policy over planning cycles on a curvy test segment.}
              \label{fig:curvypolicydist}
            \end{subfigure}
            \hspace{2mm}
            \begin{subfigure}[t]{0.40\textwidth}
              \centering
              \includegraphics{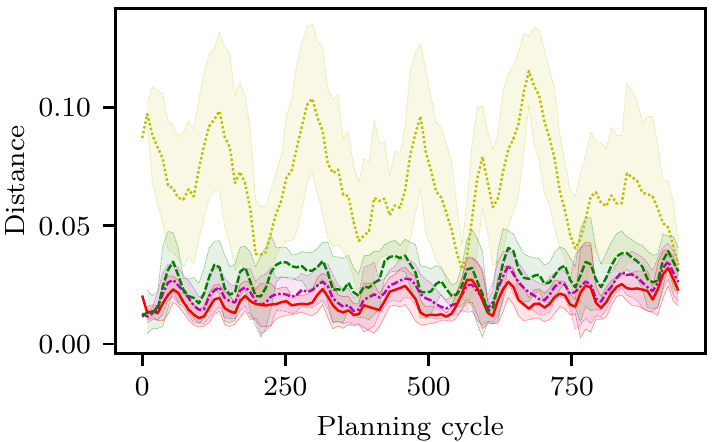}
              \caption{Distances of the optimal policy over planning cycles on a straight test segment.}
              \vspace{2mm}
              \label{fig:straightpolicydist}
            \end{subfigure}
            \begin{subfigure}[t]{0.40\textwidth}
              \centering
              \includegraphics{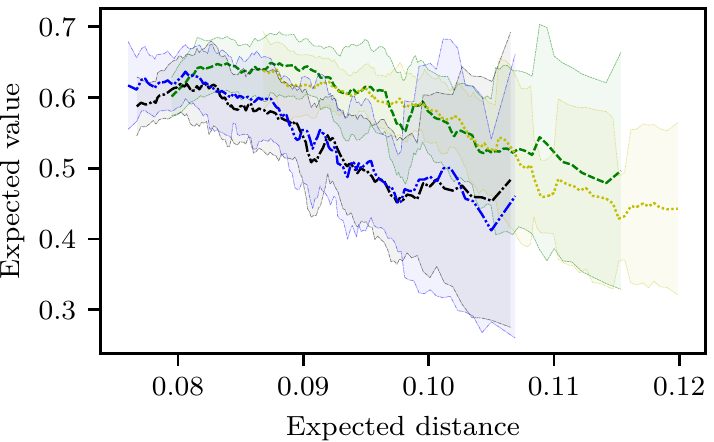}
              \caption{Expected value and distance under the expert tuned reward function of every planning cycle on a curvy test segment.}
              \label{fig:curvyexpectedvalueexpecteddistance}
            \end{subfigure}
            \hspace{2mm}
            \begin{subfigure}[t]{0.40\textwidth}
              \centering
              \includegraphics{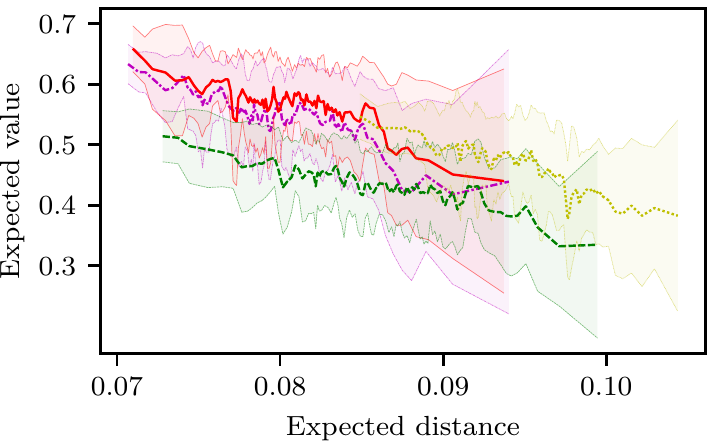}
              \caption{Expected value and distance under the expert tuned reward function of every planning cycle on a straight test segment.}
              \label{fig:straightexpectedvalueexpecteddistance}
            \end{subfigure}
            \caption{
              The tests contrast the driving style of random, learned, and expert-tuned reward functions.
              The graphs present the results of independent playbacks on a dedicated test track.
              The probability is calculated independently for every planning cycle of the MPC, whereas the policy set includes on average 4000 policies.
            }
   \label{fig:test}
\end{figure*}

In this part of the evaluation, we compare the driving style of the random, learned, and expert-tuned reward functions shown in Fig.~\ref{fig:allweights} to manual human driving.
The parameters of the reward functions allow for introspection and reasoning about the segment-specific preference.
The reward weight is inversely proportional to the preference of that feature value in the policy.
Learned reward functions are of two types:
\begin{enumerate}[(a)]
\item IRL with random initialization, hereby referred as IRL(random). Both the training trajectory set and the learning task are randomly initialized.
\item IRL with expert initialization, hereby referred as IRL(expert). Both the training trajectory set and the learning task are initialized by expert tuning.
\end{enumerate}
Using these reward functions, we run our planning algorithm on dedicated test route segments to verify the generalized performance of the optimal policies.
We carry out multiple drives over the test segments to generate representative statistics.
Fig.~\ref{fig:curvydisthist} and Fig.~\ref{fig:straightdisthist} present the projection metric distribution, which is the distance of the optimal policy to the odometry of a manual human drive for every planning cycle.
We fit a Gaussian distribution over the histogram with 200 bins of size 0.001 with 944 planning cycles for the straight and 369 planning cycles for the curvy segment.
The learned reward functions improve the driving style on all segments even in the case of random initialization.
Our evaluation metric, which is the mean distance of the optimal policy to the odometry, decreases for IRLStraight(random) by 73\% and for IRLCurve(random) by 43\%.
In case of expert-tuned initialization, IRLStraight(expert) decreased by 22\% and IRLCurve(expert) by 4\%.
The strong learning outcome in the straight segment can be attributed to the easier learning task as compared to the curvy segment.
Even though the expert-tuned reward functions do not improve substantially in terms of mean distance, they show a lower variance in distance of the optimal policy to the odometry over planning cycles after training as is shown in Fig.~\ref{fig:curvypolicydist} and Fig.~\ref{fig:straightpolicydist}.
Here we indicate variance in the distance of the optimal policy over planning cycles by one standard deviation.
The variance reduction of learned reward function depicts higher stability over planning cycles.
Hence, we are able to encode the human driving style through IRL without applying prior domain knowledge as done by motion planning experts.

Fig.~\ref{fig:curvyexpectedvalueexpecteddistance} and Fig.~\ref{fig:straightexpectedvalueexpecteddistance} present the expected value of our evaluated reward functions $r^A$ under the expert-tuned reward function $r^E$, given by ${\mathbb{E}[V(\Pi)]= \sum_{\pi \in \Pi} {p(V^{\pi}(r^A)) V^{\pi}{(r^E)}}}$.
The overall trend indicates an inverse relationship between expected value and expected distance.
The learned reward functions have lower expected distance as compared to expert tuned and random reward functions, while having a higher rate of value reduction with increasing expected distance. 
This ensures that the learned reward functions induce a high degree of bias in the policy evaluation such that the humanlike demonstrated behavior is preferred. 

\section{Conclusion and Future Work}

We utilize \textit{path integral maximum entropy} IRL to learn reward functions of a general-purpose planning algorithm.
Our method integrates well with model-based planning algorithms and allows for automated tuning of the reward function encoding the humanlike driving style.
This integration makes maximum entropy IRL tractable for high dimensional state spaces.
The automated tuning process allows us to learn reward functions for specific driving situations.
Our experiments show that learned reward functions improve the driving style exceeding the level of manual expert-tuned reward functions.
Furthermore, our approach does not require prior knowledge except the defined features of the linear reward function.
In the future, we plan to extend our IRL approach to update the reward function dynamically.

\bibliographystyle{IEEEtran}

\end{document}